\def\year{2021}\relax
\documentclass[letterpaper]{article} %
\usepackage{aaai21}  %
\usepackage{times}  %
\usepackage{helvet} %
\usepackage{courier}  %
\usepackage[hyphens]{url}  %
\usepackage{graphicx} %
\usepackage{natbib}  %
\usepackage{caption} %
\usepackage{comment}

\usepackage{algpseudocode}
\algblockdefx[parallel]{ParFor}{EndPar}[1][]{$\textbf{parallel for}$ #1 $\textbf{do}$}{$\textbf{end parallel}$}
\algrenewcommand{\alglinenumber}[1]{\fontsize{6.5}{7}\selectfont#1}
\algtext*{EndPar}

\algblockdefx[parallel]{parfor}{endpar}[1][]{$\textbf{parallel for}$ #1 $\textbf{do}$}{$\textbf{end parallel}$}
\algrenewcommand{\alglinenumber}[1]{\scriptsize#1:}

\algblockdefx[parallel]{parallelfor}{parallelend}
[1][]{\textbf{parallel for} #1}
{\textbf{end parallel}}

\usepackage{nicefrac}

\algnotext{EndFor}
\algnotext{EndIf}
\algnotext{EndProcedure}

\DeclareMathAlphabet\mathbfcal{OMS}{cmsy}{b}{n}

\usepackage{booktabs} %
\usepackage{epstopdf}
\usepackage{comment}
\usepackage{tabularx}
\usepackage{paralist}
\usepackage{balance}
\usepackage{listings}
\usepackage{subcaption}
\usepackage{float}
\usepackage[ruled,vlined]{algorithm2e}
\usepackage[super]{nth}
\usepackage[switch]{lineno} %

\usepackage{xcolor,colortbl}
\definecolor{thedarkblue}{RGB}{0,0,120} %
\definecolor{mydarkblue}{rgb}{0,0.08,0.45} %
\definecolor{darkblue}{rgb}{0,0.08,180}
\colorlet{TufteRed}{red!80!black}

\definecolor{theblue}{RGB}{0,0,180}
\colorlet{thered}{TufteRed}

\usepackage{hyperref}
\hypersetup{%
    colorlinks=true,
    linkcolor=black,
    citecolor=black,
    filecolor=black,
    urlcolor=black}

\usepackage{microtype}
\usepackage{balance}

\usepackage{amsmath,amssymb,amsthm}
\usepackage{dsfont}

\newcommand{\eat}[1]{\ignorespaces}

\usepackage{tikz}
\usepackage{verbatim}
\usetikzlibrary{arrows}
\usetikzlibrary{shapes,snakes}
\usetikzlibrary{decorations.pathmorphing} %
\usetikzlibrary{fit}					%
\usetikzlibrary{backgrounds}	%

\usepackage{ragged2e}
\usepackage{multirow}
\usepackage{microtype}
\usepackage{balance}
\usepackage{setspace}

\graphicspath{{./}{./graphics/}}
\newcolumntype{H}{>{\setbox0=\hbox\bgroup}c<{\egroup}@{}}

\newcolumntype{R}[1]{>{\RaggedLeft\arraybackslash}} %
\newcolumntype{L}[1]{>{\RaggedRight\arraybackslash}} %

\newcommand{\abs}[1]{\left|#1\right|}

\newcommand{\eg}{\emph{e.g.}}

\newtheorem{theorem}{\bfseries{Theorem}}

\newtheorem{Definition}{\bfseries{Definition}}

\providecommand{\mat}[1]{\boldsymbol{\mathrm{#1}}}%
\renewcommand{\vec}[1]{\boldsymbol{\mathrm{#1}}}

\DeclareMathOperator{\hugeE}{\mbox{\huge\raise-0.3ex\hbox{E}}}
\DeclareMathOperator{\p}{\mathbb{P}}
\DeclareMathOperator{\hugep}{\mbox{\huge\raise-0.3ex\hbox{$\p$}}}

\providecommand{\graph}{\mathcal}

\newcommand{\RR}{\mathbb{R}}

\providecommand{\eye}{\mat{I}}
\providecommand{\mA}{\ensuremath{\mat{A}}}

\providecommand{\mC}{\ensuremath{\mat{C}}}
\providecommand{\mD}{\ensuremath{\mat{D}}}
\providecommand{\mE}{\ensuremath{\mat{E}}}

\providecommand{\mH}{\ensuremath{\mat{H}}}

\providecommand{\mY}{\ensuremath{\mat{Y}}}

\providecommand{\ve}{\ensuremath{\vec{e}}}

\newcommand{\V}[1]{\mathbf{#1}}
\newcommand{\R}{\mathbb{R}}

\newcommand{\vertexSet}{\mathcal{V}}
\newcommand{\edgeSet}{\mathcal{E}}

\newcommand{\matA}{\mathbf{A}}

\newcommand{\matD}{\mathbf{D}}

\newcommand{\matX}{\mathbf{X}}

\newcommand{\matW}{\mathbf{W}}
\newcommand{\matH}{\mathbf{H}}
\newcommand{\matI}{\mathbf{I}}
\newcommand{\matL}{\mathbf{L}}

\newcommand{\vecy}{\mathbf{y}}

\newcommand{\setY}{\mathcal{Y}}
\newcommand{\setT}{\mathcal{T}_{\vertexSet}}

\newcommand{\method}{CPGNN\xspace}
\newcommand{\methodbeyond}{H${_2}$GCN}

\newcommand{\norm}[2]{\Vert{#1}\Vert_{#2}}

\DeclareMathOperator{\sinkhornknopp}{Sinkhorn-Knopp}

\title{Graph Neural Networks with Heterophily}
\author{
    Jiong Zhu,\textsuperscript{\rm 1} Ryan A. Rossi,\textsuperscript{\rm 2} Anup Rao,\textsuperscript{\rm 2} Tung Mai,\textsuperscript{\rm 2} \\
    Nedim Lipka,\textsuperscript{\rm 2} Nesreen K. Ahmed,\textsuperscript{\rm 3} Danai Koutra\textsuperscript{\rm 1}
}

\affiliations{

    \textsuperscript{\rm 1}University of Michigan, Ann Arbor, USA\\
    \textsuperscript{\rm 2}Adobe Research, San Jose, USA \\
    \textsuperscript{\rm 3}Intel Labs, Santa Clara, USA \\

    jiongzhu@umich.edu, \{ryrossi, anuprao, tumai, lipka\}@adobe.com, nesreen.k.ahmed@intel.com, dkoutra@umich.edu %
}

\begin{document}

\maketitle

\begin{abstract}
Graph Neural Networks (GNNs) have proven to be useful for many different practical applications. 
However, many existing GNN models have implicitly assumed homophily among the nodes connected in the graph, and therefore have largely overlooked
the important setting of heterophily, 
where most connected nodes are from different classes.
In this work, we propose a novel framework called \method{} that generalizes GNNs for graphs with either homophily or heterophily. 
The proposed framework incorporates an interpretable compatibility matrix for modeling the heterophily or homophily level in the graph, which can be learned in an end-to-end fashion, enabling it to go beyond the assumption of strong homophily. 
Theoretically, we show that replacing the compatibility matrix in our framework with the identity (which represents pure homophily) reduces to GCN. 
Our extensive experiments demonstrate the effectiveness of our approach in more realistic and challenging experimental settings with significantly less training data compared to previous works: CPGNN variants achieve state-of-the-art results in heterophily settings with or without contextual node features, while maintaining comparable performance in homophily settings. 
\end{abstract}

\section{Introduction}

As a powerful approach for learning and extracting information from relational data, Graph Neural Network (GNN) models have gained wide research interest~\cite{scarselli2008graph} and been adapted in applications including recommendation systems~\cite{ying2018graph}, bioinformatics~\cite{zitnik2018modeling, yan2019groupinn}, fraud detection~\cite{dou2020enhancing}, and more. While many different GNN models have been proposed, existing methods have largely overlooked several limitations in their formulations: (1) \emph{implicit homophily assumptions}; (2) \emph{heavy reliance on contextual node features}. First, many GNN models, including the most popular GNN variant proposed by \citet{kipf2017semi}, implicitly assume \emph{homophily} in the graph, where most connections happen among nodes in the same class or with alike features~\cite{mcpherson2001birds}. 
This assumption has affected the design of many GNN models, which tend to generate similar representations for nodes within close proximity, as studied in previous works~\cite{role2vec-ijcai18,from-comm-to-structural-role-embeddings, wu2019simplifying}. 
However, there are also many instances in the real world where nodes of different classes are more likely to connect with one another --- in idiom, this phenomenon can be described as ``opposites attract''.
As we observe empirically, many GNN models which are designed under implicit homophily assumptions suffer from poor performance in heterophily settings, which can be problematic for applications like fraudster detection~\cite{pandit2007netprobe,dou2020enhancing} or analysis of protein structures~\cite{fout2017protein}, 
where heterophilous connections are common.
Second, many existing models rely heavily on contextual node features to derive intermediate representations of each node, which is then propagated within the graph. 
While in a few networks like citation networks, node features are able to provide powerful node-level contextual information for downstream applications, in more common cases the contextual information are largely missing, insufficient or incomplete, which can significantly degrade the performance for some models. 
Moreover, complex transformation of input features usually requires the model to adopt a large number of learnable parameters, which need more data and computational resources to train and are hard to interpret.

\begin{figure*}[t!]
    \centering
    \includegraphics[width=.95\textwidth]{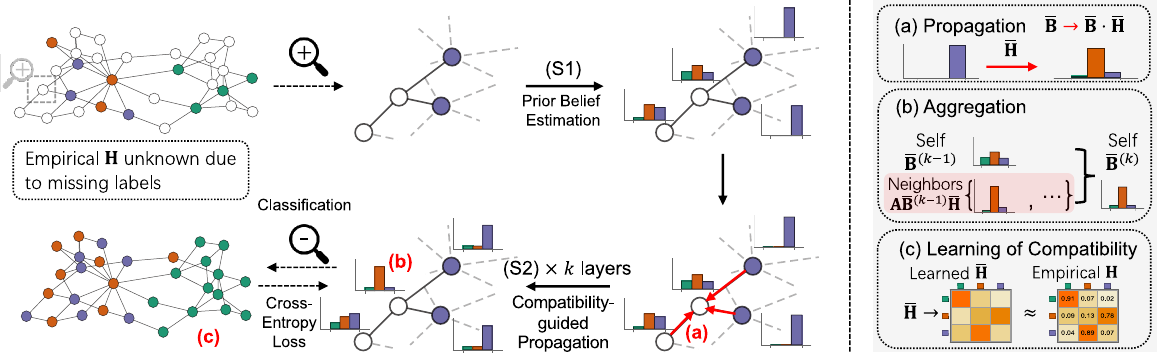}
    \caption{The general pipeline of the proposed framework (\method{}) with $k$ propagation layers (\S\ref{sec:framework-design}). As an example, we use a graph with mixed homophily and heterophily, with node colors representing class labels: nodes in green show strong \emph{homophily}, while nodes in orange and purple show strong \emph{heterophily}. \method{} framework first generates prior belief estimations using an off-the-shelf neural network classifier, which utilizes node features if available (S1). 
    The prior beliefs are then propagated within their neighborhoods guided by the learned compatibility matrix, and each node aggregates beliefs sent from its neighbors to update its own beliefs (S2). 
    {
    We describe the backward training process, including end-to-end learning of compatibility in \S\ref{sec:training-procedure}.}}
    \label{fig:method-pipeline}
    \vspace{-0.3cm}
\end{figure*}

In this work, we propose \method{},
a novel approach that incorporates into GNNs a 
compatibility matrix that captures both heterophily and homophily by modeling the likelihood of connections between nodes in different classes.
This novel design overcomes the drawbacks of existing GNNs mentioned above: 
it enables GNNs to appropriately learn from graphs with either homophily \emph{or} heterophily, and is able to achieve satisfactory performance even in the cases of missing and incomplete node features. Moreover, the end-to-end learning of the class compatibility matrix effectively recovers the ground-truth underlying compatibility information, which is hard to infer from limited training data, and provides insights for understanding the connectivity patterns within the graph.
Finally, the key idea proposed in this work can naturally be used to generalize many other GNN-based methods by incorporating and learning the heterophily compatibility matrix $\mH$ in a similar fashion.

\medskip\noindent
We summarize the main contributions as follows: 
\begin{itemize}
\item \textbf{Heterophily Generalization of GNNs}.
We describe a generalization of GNNs to heterophily settings by incorporating a compatibility matrix $\mH$ into GNN-based methods, which is learned in an end-to-end fashion.

    \item \textbf{\method{} Framework}. 
    We propose \method{}, a novel approach that directly models and learns the class compatibility matrix $\mH$ in GNN-based methods.
    This formulation gives rise to many advantages including better 
    effectiveness for graphs with either homophily or heterophily, and for graphs with or without node features.
    We release \method{} at {\url{https://github.com/GemsLab/CPGNN}}.

    \item \textbf{Comprehensive Evaluation}.
    We conduct extensive experiments to compare the performance of \method{} with baseline methods under a more \emph{realistic} experimental setup by using significantly fewer training data comparing to the few previous works which address heterophily~\cite{pei2019geom,zhu2020beyond}.
    These experiments demonstrate the effectiveness of incorporating the compatibility matrix $\mH$ into GNN-based methods. 
\end{itemize}

\section{Framework}

In this section we introduce our \method{} framework, after presenting the problem setup and important definitions.

\subsection{Preliminaries}

\noindent \textbf{Problem Setup.} We focus on the problem of semi-supervised node classification on a simple graph $\graph{G} = (\vertexSet, \edgeSet)$, where $\vertexSet$ and $\edgeSet$ are the node- and edge-sets respectively, and $\setY$ is the set of possible class labels 
for $v \in \vertexSet$. 
Given a training set $\setT \subset \vertexSet$ with known class labels $y_v$ for all $v \in \setT$, and (optionally) contextual feature vectors $\Vec{x}_v$ for $v \in \vertexSet$, we aim to infer the unknown class labels $y_u$ for all $u \in (\vertexSet - \setT)$. 
For subsequent discussions, we use $\matA \in \{0,1\}^{|\vertexSet| \times |\vertexSet|}$ for the adjacency matrix \emph{with self-loops removed}, $\vecy \in \setY^{|\vertexSet|}$ as the ground-truth class label vector for all nodes, and $\matX \in \RR^{|\vertexSet| \times F} $ for the node feature matrix.

\vspace{0.1cm}
\noindent \textbf{Definitions.} We now introduce two key concepts for modeling the homophily level in the graph with respect to the class labels: (1) \emph{homophily ratio}, and (2) \emph{compatibility matrix}. 
\begin{Definition}[\bf Homophily Ratio $h$]
Let $\mC \in \RR^{|\setY| \times |\setY|}$ where $C_{ij} = |\{(u, v): (u, v) \in \edgeSet \wedge y_u = i \wedge y_v = j\}|$, $\mD = \mathrm{diag}(\{C_{ii}: i=1,\ldots,|\setY|\})$, and $\ve \in \RR^{|\vertexSet|}$ be an all-ones vector.
The homophily ratio is defined as 
$h = \frac{\ve^{\top} \mD \ve}{\;\ve^{\top} \mC \ve\;}$.
\end{Definition}
The homophily ratio $h$ defined above is good for measuring the \emph{overall} homophily level in the graph. By definition, we have $h\in [0, 1]$: graphs with $h$ closer to $1$ tend to have more edges connecting nodes within the same class, or \emph{stronger homophily}; on the other hand, graphs with $h$ closer to 0 have more edges connecting nodes in different classes, or a \emph{stronger heterophily}. 
However, the actual homophily level is not necessarily uniform within all parts of the graph. One common case is that the homophily level varies among different pairs of classes, where it is more likely for nodes between some pair of classes to connect than some other pairs. To measure the variability of the homophily level, we define the \emph{compatibility matrix} $\mH$ as \citet{zhu2020beyond} as follows: 

\begin{Definition}[\bf Compatibility Matrix $\mH$]
\label{dfn:compatibility-matrix}
Let $\mY \in \RR^{|\vertexSet| \times |\setY|}$ where $Y_{vj}=1$ if $y_v = j$, and $Y_{vj}=0$ otherwise. Then, the compatibility matrix $\mH$ is defined as: 
\begin{equation}\label{eq:heterophily-matrix-formulation} %
\small
\mH = (\mY^{\top}\mA\mY) \oslash (\mY^{\top}\mA\mE)
\end{equation}\noindent
where $\oslash$ is Hadamard (element-wise) division and 
$\mE$ is a $|\vertexSet|\times |\setY|$ all-ones matrix.
\end{Definition}
In node classification settings, compatibility matrix $\mH$ models the (empirical) probability of nodes belonging to each pair of classes to connect. 
More generally, $\mH$ can be used to model any discrete attribute;
in that case, $\mH_{ij}$ is the probability that a node with attribute value $i$ connects with a node with value $j$.
Modeling $\mH$ in GNNs is beneficial for heterophily settings, but calculating the exact $\mH$ would require knowledge to the class labels of all nodes in the graph, which violates the semi-supervised node classification setting. 
Therefore, it is not possible to incorporate exact $\mH$ into graph neural networks. In the following sections, we propose \method{}, which is capable of learning $\mH$ in an end-to-end way based on a rough initial estimation. 

\subsection{Framework Design}
\label{sec:framework-design}
The \method{} framework consists of two stages: 
(S1) prior belief estimation; and 
(S2) compatibility-guided propagation. 
We visualize the \method{} framework in Fig.~\ref{fig:method-pipeline}.

\vspace{0.1cm}
\noindent \textbf{(S1) Prior Belief Estimation} The goal for the first step is to estimate per node $v \in \vertexSet$ a prior belief $\mathbf{b}_v \in \R^{|\setY|}$ of its class label $y_v \in \setY$ from the node features $\matX$. 
{This separate, explicit prior belief estimation stage enables the use of any off-the-shelf neural network classifier as the estimator, and thus can accommodate different types of node features.}
In this work, we consider the following models as the estimators:
\begin{itemize}
    \item \texttt{MLP}, a graph-agnostic multi-layer perceptron. Specifically, the $k$-th layer of the MLP can be formulated as following:
    \begin{equation}
        \label{eq:prior-estimator-mlp}
        \small
        \V{R}^{(k)} = \sigma(\V{R}^{(k-1)} \matW^{(k)}),
    \end{equation}
    where $\matW^{(k)}$ are learnable parameters, and $\V{R}^{(0)} = \matX$. We call our method with MLP-based estimator \texttt{\method-MLP}.
    \item \texttt{GCN-Cheby}~\cite{defferrard2016convolutional}. %
    We instantiate the model using a \nth{2}-order Chebyshev polynomial, where the $k$-th layer is parameterized as:
    \begin{equation}
    \small
       \textstyle \mathbf{R}^{(k)} = \sigma\Big(\sum_{i=0}^{2} T_i(\tilde{\matL}) \V{R}^{(k-1)} \matW^{(k)}_i \Big).
    \end{equation}
    $\matW_i^{(k)}$ are learnable parameters, $\V{R}^{(0)} = \matX$, and $T_i(\tilde{\matL})$ is the $i$-th order of the Chebyshev polynomial of $\tilde{\matL} = \matL - \matI$ defined recursively as: 
    $$T_i(\tilde{\matL}) = 2 \tilde{\matL} T_{i-1}(\tilde{\matL}) - T_{i-2}(\tilde{\matL})$$
    with $T_0(\tilde{\matL}) = \matI$ and $T_1(\tilde{\matL}) = \tilde{\matL} =  -\matD^{-\frac{1}{2}}\matA\matD^{-\frac{1}{2}}$. We refer to our Cheby-based method as \texttt{\method-Cheby}.
\end{itemize}
We note that the performance of CPGNN is affected by the choice of the estimator. It is important to choose an estimator that is \emph{not} constrained by the homophily assumption (e.g., our above-mentioned choices), so that it does not hinder the performance in heterophilous graphs.

Denote the output of the final layer of the estimator as $\V{R}^{(K)}$, then the prior belief $\mathbf{B}_p$ of nodes can be given as
\begin{equation}
\small
    \label{eq:prior-belief}
    \mathbf{B}_p = \mathrm{softmax}(\V{R}^{(K)})
\end{equation}
To facilitate subsequent discussions, we denote the trainable parameters of a general prior belief estimator as $\mathbf{\Theta}_p$, and the prior belief of node $v$ derived by the estimator as $\mathcal{B}_p(v; \mathbf{\Theta}_p)$. 

\vspace{0.1cm}
\noindent \textbf{(S2) Compatibility-guided Propagation} 
We propagate the prior beliefs of nodes within their neighborhoods using a parameterized, end-to-end trainable compatibility matrix $\bar{\mathbf{H}}$.

To propagate the belief vectors through linear formulations, 
following \citet{gatterbauer2015linearized}, 
we first center $\mathbf{B}_p$ with
\begin{equation}
\small 
    \bar{\mathbf{B}}^{(0)} = \mathbf{B}_p - \tfrac{1}{|\setY|}
\end{equation}
We parameterize the compatibility matrix as $\bar{\mH}$ 
to replace the weight matrix $\matW$ in traditional GNN models as the end-to-end trainable parameter.
We formulate each layer as: 
\begin{equation}
\small 
    \label{eq:method-basic-form}
    \bar{\mathbf{B}}^{(k)} = \bar{\mathbf{B}}^{(0)} + \matA \bar{\mathbf{B}}^{(k-1)} \bar{\mH}
\end{equation}
{Each layer propagates and updates the current belief per node in its neighborhood.
After $K$ layers, we have the final belief}
\begin{equation}
    \V{B}_f = \mathrm{softmax}(\bar{\mathbf{B}}^{(K)}).
\end{equation}
We similarly denote $\mathcal{B}_f(v; \bar{\mH}, \mathbf{\Theta}_p)$ as the final belief for node $v$, where parameters $\mathbf{\Theta}_p$ are from the prior belief estimator.

\subsection{Training Procedure}
\label{sec:training-procedure}
\paragraph{Pretraining of Prior Belief Estimator.} 
We pretrain the prior belief estimator for $\beta_1$ iterations so that $\bar{\mathbf{H}}$ can then be better initialized with informative prior beliefs. Specifically, the pretraining process aims to minimize the loss function
\begin{equation}
\small 
   \textstyle \mathcal{L}_p(\mathbf{\Theta}_p) = \sum_{v \in \setT} \mathcal{H}\left(\mathcal{B}_p(v; \mathbf{\Theta}_p), y_v\right) + \lambda_p \norm{\mathbf{\Theta}_p}{2},
\end{equation}
where $\mathcal{H}$ corresponds to the cross entropy, and $\lambda_p$ is the L2 regularization weight for the prior belief estimator. 
Through an ablation study (App. \S\ref{sec:app-detailed_results}, Fig.~\ref{fig:design-ablations-extra}), we show that pretraining prior belief estimator helps increase the final performance. 

\vspace{0.1cm}
\noindent \textbf{Initialization and Regularization of $\,\bar{\matH}$.} 
We empirically found that initializing the parameters $\bar{\matH}$ with an estimation of the unknown compatibility matrix $\matH$ can lead to better performance (cf. \S\ref{sec:exp-ablation-study}, Fig. \ref{fig:design-ablation-1}). We derive the estimation using node labels in the training set $\setT$, and prior belief $\mathbf{B}_p$ estimated in Eq. \eqref{eq:prior-belief} after the pretraining stage. More specifically, denote the training mask matrix $\mathbf{M}$ as: 
\begin{equation}
\small 
    [\mathbf{M}]_{i, :} = 
    \left\{
    \begin{array}{ll}
        \mathbf{1},\;\text{if }i \in \setT \\
        \mathbf{0},\;\text{otherwise}
    \end{array}
    \right.
\end{equation}
and the enhanced belief matrix $\tilde{\mathbf{B}}$, which makes use of known node labels $\mathbf{Y}_{\mathrm{train}} = \mathbf{M} \circ \mathbf{Y}$ in the training set $\setT$, as
\begin{equation}
\small 
    \tilde{\mathbf{B}} = \mathbf{Y}_{\mathrm{train}} + (1 - \mathbf{M}) \circ \mathbf{B}_p
\end{equation}
where $\circ$ is the element-wise product. The estimation $\hat{\matH}$ of the unknown compatibility matrix $\matH$ is derived as 
\begin{equation}
\small
    \label{eq:initial-H-estimation}
    \hat{\matH} = \sinkhornknopp\left(\mathbf{Y}_{\mathrm{train}}^\top \matA \tilde{\mathbf{B}}\right)
\end{equation}
where the use of the \citet{sinkhorn1967concerning} algorithm is to ensure that $\hat{\matH}$ is doubly stochastic. 
We find that a doubly-stochastic and symmetric initial value for $\bar{\matH}$ boosts the training when using multiple propagation layers. 
Thus, we initialize the parameter $\bar{\matH}$ as $\bar{\mathbf{H}}_0 = \tfrac{1}{2} (\hat{\matH} + \hat{\matH}^\top) - \tfrac{1}{|\setY|}$, where $\hat{\matH}$ is centered around 0 (similar to $\mathbf{B}_p$). 
To ensure the rows of $\bar{\matH}$ remain centered around 0 throughout the training process, we adopt the following regularization term $\Phi(\bar{\mH})$ for $\bar{\matH}$: 
{
\begin{equation}
\small 
    \label{eq:H-zero-reg}
   \textstyle \Phi(\bar{\mH}) = \eta \sum_i \left| \sum_j \bar{\matH}_{ij} \right|
\end{equation}
where $\eta$ is the regularization weight for $\bar{\matH}$.
}

\vspace{0.1cm}
\noindent \textbf{Loss Function for \method{} Training.} Putting everything together, we obtain the loss function for training \method{}: 
{
\begin{equation}
    \footnotesize
    \label{eq:overall-loss}
    \mathcal{L}_f (\bar{\mathbf{H}}, \mathbf{\Theta}_p) = 
    \sum_{v \in \setT} \mathcal{H}\left(\mathcal{B}_f(v; \bar{\mH}, \mathbf{\Theta}_p),y_v\right)
    + \lambda_p \norm{\mathbf{\Theta}_p}{2} + \Phi(\bar{\mH}) 
\end{equation}
}%
The loss function consists of three parts: (1) the cross entropy loss from the \method{} output; (2) the {L2 regularization} from the prior belief estimator; and (3) a regularization term that keeps $\bar{\matH}$ centered around 0 throughout the training process. The {last term is} novel for the \method{} formulation, and helps increase the performance of \method{}, as we show later through an ablation study (\S\ref{sec:exp-ablation-study}). %

\subsection{Interpretation of Parameters $\bar{\mH}$}
\label{sec:h-matrix-estimation}

Unlike the hard-to-interpret weight matrix $\matW$ in classic GNNs, parameter $\bar{\mathbf{H}}$ in CPGNN can be easily understood: it captures the probability that node pairs in specific classes connect with each other.
Through an inverse of the initialization process, we can obtain an estimation of the compatibility matrix $\hat{\mathbf{H}}$ after training from learned parameter $\bar{\mathbf{H}}$ as follows:
\begin{equation}
\small 
    \hat{\mathbf{H}} = \sinkhornknopp(\tfrac{1}{\alpha}\bar{\mathbf{H}} + \tfrac{1}{|\setY|})
\end{equation}
where $\alpha = \min \{a \geq 1: 1 \geq \tfrac{1}{a}\bar{\mathbf{H}} + \tfrac{1}{|\setY|} \geq 0\}$ is a recalibration factor ensuring that the obtained $\hat{\mathbf{H}}$ is a valid stochastic matrix. In \S\ref{sec:exp-h-estimation}, we provide an example of the estimated $\hat{\mathbf{H}}$ after training, and show the improvements in estimation error compared to the initial estimation by Eq. \eqref{eq:initial-H-estimation}.

\section{Theoretical Analysis}
\label{sec:theory}

\paragraph{Theoretical Connections.}  
Theorem~\ref{thm:sgc-equivalence} establishes the theoretical result that \method{} can be reduced to a simplified version of GCN when $\mH = \eye$. 
Intuitively, replacing $\mH$ with $\eye$ indicates a pure homophily assumption, and thus shows exactly the reason that GCN-based methods have a strong homophily assumption built-in, and therefore perform worse for graphs \emph{without} strong homophily.

\begin{theorem}
\label{thm:sgc-equivalence}
The forward pass formulation of a 1-layer SGC~\cite{wu2019simplifying}, a simplified version of GCN without the non-linearities and adjacency matrix normalization,
\begin{equation}
\small 
    \label{eq:sgc-proof-formulation}
    \V{B}_f = \mathrm{softmax}\left(\left(\mA + \eye\right) \mathbf{X \Theta}\right)
\end{equation}
where $\mathbf{\Theta}$ denotes the model parameter, can be treated as a special case of \method{} with compatibility matrix $\mH$ fixed as $\eye$ and non-linearity removed in the prior belief estimator.
\end{theorem}

\vspace{0.1cm}
\noindent\textbf{Proof} The formulation of \method{} with 1 aggregation layer can be written as follows:
\begin{equation}
\small
    \label{eq:sgc-proof-1}
    \V{B}_f = \mathrm{softmax}(\bar{\mathbf{B}}^{(1)}) = \mathrm{softmax}\left( \bar{\mathbf{B}}^{(0)} + \matA \bar{\mathbf{B}}^{(0)} \mH \right)
\end{equation}
Now consider a 1-layer MLP (Eq. \eqref{eq:prior-estimator-mlp}) as the prior belief estimator. Since we assumed that the non-linearity is removed in the prior belief estimator, we can assume that $\mathbf{B}_p$ is already centered. Therefore, we have
\begin{equation}
\small
    \label{eq:sgc-proof-2}
    \bar{\mathbf{B}}^{(0)} = \mathbf{B}_p = \V{R}^{(K)} = \V{R}^{(0)} \matW^{(0)} = \mathbf{X} \matW^{(0)}
\end{equation}
where $\matW^{(0)}$ is the trainable parameter for MLP. Plug in Eq. \eqref{eq:sgc-proof-2} into Eq. \eqref{eq:sgc-proof-1}, we have 
\begin{equation}
\small
    \V{B}_f = \mathrm{softmax}\left( \mathbf{X} \matW^{(0)} + \matA \mathbf{X} \matW^{(0)} \mH \right)
\end{equation}
Fixing compatibility matrix $\mH$ fixed as $\eye$, and we have
\begin{align}
\small
     \V{B}_f  = \mathrm{softmax}\left( (\matA+\eye) \mathbf{X} \matW^{(0)} \right)
\end{align}
As $\matW^{(0)}$ is a trainable parameter equivalent to $\mathbf{\Theta}$ in Eq. \eqref{eq:sgc-proof-formulation}, the notation is interchangeable. Thus, the simplified GCN formulation as in Eq. \eqref{eq:sgc-proof-formulation} can be reduced to a special case of \method{} with compatibility matrix $\mH =\eye$. \hfill $\blacksquare$

\vspace{0.1cm}
\noindent\textbf{Time and Space Complexity of \method{}} 
Let $|\edgeSet|$ and $|\vertexSet|$ denote the number of edges and nodes in $\mathcal{G}$, respectively.
Further, let $|\edgeSet_i|$ denote the number of node pairs in $\mathcal{G}$ within $i$-hop distance (\eg, $|\edgeSet_1| = |\edgeSet|$) and $|\setY|$ denotes the number of unique class labels.
We assume the graph adjacency matrix $\matA$ and node feature matrix $\matX$ are stored as sparse matrices.

\method{} only introduces $\mathrm{O}(|\edgeSet| |\setY|^2)$ extra time over the selected prior belief estimator in the propagation stage (S2). Therefore, the overall complexity for \method{} is largely determined by the time complexity of the selected prior belief estimator:
when using MLP as prior belief estimator (Stage S1), the overall time complexity of \method{}-MLP is $\mathrm{O}(|\edgeSet| |\setY|^2 + |\vertexSet||\setY| + \mathrm{nnz}(\matX))$,
while the overall time complexity of an $\alpha$-order \method{}-Cheby is
$\mathrm{O}(|\edgeSet| |\setY|^2 + |\vertexSet||\setY| + \mathrm{nnz}(\matX) + |\edgeSet_{\alpha-1}|d_{\mathrm{max}} + |\edgeSet_{\alpha}|)$, where $d_{\mathrm{max}}$ is the max degree of a node in $\mathcal{G}$ and $\mathrm{nnz}(\matX)$ is the number of nonzeros in $\matX$.

The overall space complexity of \method{} is $\mathrm{O}(|\edgeSet| + |\vertexSet| |\setY| + |\setY|^2 + \mathrm{nnz}(\matX))$, which also takes into account the space complexity for the two discussed prior belief estimators above (MLP and GCN-Cheby).

\begin{table}[b!]
	\centering
	\vspace{-0.2cm}
    \small
	\begin{tabular}{lrrcccHH}
		\toprule 
		& \multicolumn{1}{c}{\textbf{ $|\vertexSet|$}}  & \multicolumn{1}{c}{\textbf{$|\edgeSet|$}} & \multicolumn{1}{c}{\textbf{$|\setY|$}} & %
		\multicolumn{1}{c}{\textbf{$F$}} & 
		\multicolumn{1}{c}{\textbf{$h$}} & 
		 \\ \toprule %
	\textbf{syn-} & \multirow{2}{*}{10,000} & 59,640--  & \multirow{2}{*}{10} & %
	\multirow{2}{*}{100} & [0, 0.1,  & $ 33 $ \\ 
	\textbf{products} &  &  59,648 &  & %
	 & \ldots, 1] & (3 per $h$)\\
	\midrule
	{\bf Texas} & 183 & 295 & 5 & 1703 & 0.11 & 1\\
	{\bf Squirrel} & 5,201 & 198,493 & 5 & 2,089 & 0.22 & 1 \\
	{\bf Chameleon} & 2,277 & 31,421 & 5 & 2,325 & 0.23 & 1\\
	{\bf CiteSeer} & 3,327 & 4,676 & 6 & 3,703 & 0.74 & 1\\
	{\bf Pubmed} & 19,717 & 44,327 & 3 & 500 & 0.8 & 1\\
	{\bf Cora} & 2,708 & 5,278 & 7 & 1,433 & 0.81 & 1\\ \bottomrule

	\end{tabular}
 \caption{Statistics for our synthetic and real graphs: number of nodes, edges, classes, and features, and homophily ratio.} 
 \label{tab:synthetic-stats}
\end{table}

\section{Experiments}\label{sec:exp}
We design experiments to investigate the effectiveness of the proposed framework for node classification with \emph{and} without contextual features using both synthetic and real-world graphs with heterophily and strong homophily.

\subsection{Methods and Datasets}

\vspace{0.1cm}
\noindent \textbf{Methods.}
We test the two formulations discussed in \S\ref{sec:framework-design}: \texttt{CPGNN-\-MLP} and \texttt{CPGNN-\-Cheby}. Each formulation is tested with either 1 or 2 aggregation layers, leading to 4 variants in total. We compared our methods with the following baselines,
some of which are reported to be competitive under heterophily~\cite{zhu2020beyond}: GCN~\cite{kipf2017semi}, GAT~\cite{velivckovic2017graph}, GCN-Cheby~\cite{defferrard2016convolutional, kipf2017semi}, GraphSAGE~\cite{hamilton2017inductive}, MixHop~\cite{MixHop}, 
and \methodbeyond~\cite{zhu2020beyond}.
We also consider MLP as a graph-agnostic baseline. 
We provide hardware and software specifications and details on hyperparameter tuning in App. \ref{sec:app-experimental-specs} and \ref{sec:app-hyperparameter}.

\vspace{0.1cm}
\noindent \textbf{Datasets.}
We investigate \method{} using both synthetic and real-world graphs. 
For synthetic benchmarks, we generate graphs and node labels following an approach similar to \citet{karimi2017visibility17} and \citet{MixHop}, which expands the Barab\'{a}si-Albert model with configurable class compatibility settings. We assign to the nodes feature vectors from 
the recently announced Open Graph Benchmark~\cite{hu2020ogb}, which includes only graphs with homophily.
We detail the algorithms for generating synthetic benchmarks in App.~\ref{sec:app-synthetic-graph}. 
For real-world graph data, we consider graphs with heterophily and homophily. 
We use 3 heterophilous graphs, namely Texas, Squirrel and Chameleon~\cite{rozemberczki2019multiscale}, and 3 widely adopted graphs with strong homophily, which are Cora, Pubmed and Citeseer~\cite{sen2008collective,namata2012query}. We use the features and class labels provided by \citet{pei2019geom}.

\subsection{Node Classification \emph{with} Contextual Features} \label{sec:exp-classification-with-contextual-features}

\vspace{0.1cm}
\noindent \textbf{Experimental Setup.}
For synthetic experiments, we generate 3 synthetic graphs for every heterophily level $h \in \{0,0.1,0.2,\ldots,0.9,1\}$.
We then randomly select 10\% of nodes in each class for training, 10\% for validation, and 80\% for testing, and report the average classification accuracy as performance of each model on all instances with the same level of heterophily.
Using synthetic graphs for evaluation enables us to better understand how the model performance changes as a function of the level of heterophily in the graph.
Hence, we vary the level of heterophily in the graph going from strong heterophily all the way to strong homophily while holding other factors constant such as degree distribution and differences in contextual features.
On real-world graphs, we generate 10 random splits for training, validation and test sets; for each split we randomly select 10\% of nodes in each class to form the training set, with another 10\% for the validation set and the remaining as the test set. 
Notice that we are using a significantly smaller fraction of training samples compared to previous works that address heterophily~\cite{pei2019geom,zhu2020beyond}. 
This is a more realistic assumption in many real-world applications.

\begin{figure}[t]
    \centering
    \includegraphics[keepaspectratio, width=.75\columnwidth,trim={0 0 2cm 2.6cm},clip]{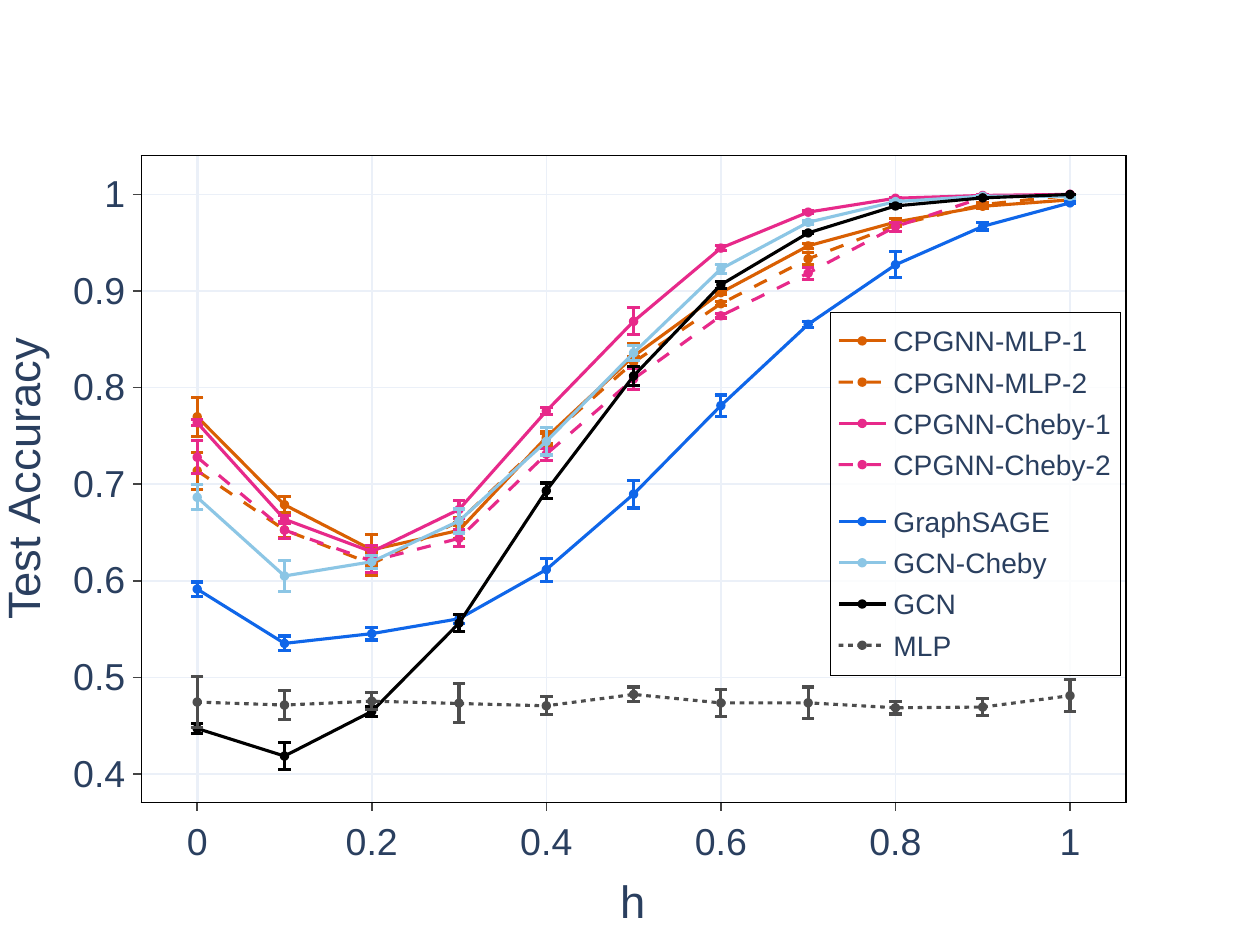}
    \vspace{-3mm}
    \caption{Mean classification accuracy of \method{} and baselines on synthetic benchmark \texttt{syn-products} 
    (cf. Table \ref{table:syn-products-table-10-est-h} for detailed results). 
    }
    \label{fig:syn-benchmark-results}
    \vspace{-0.3cm}
\end{figure}

\vspace{0.1cm}
\noindent \textbf{Synthetic Benchmarks.} 
We compare the performance of \method{} to the state-of-the-art methods in Fig. \ref{fig:syn-benchmark-results}.
Notably, we observe that \method{}-Cheby-1 consistently outperforms the baseline methods across the full spectrum of low to high homophily (or high to low heterophily).
Furthermore, compared to our \method{} variants, it performs the best in all settings with $h \geq 0.3$.
For $h < 0.3$, \method{}-MLP-1 outperforms it, and in fact performs the best overall for graphs with strong heterophily. 
More importantly, \method{} has a significant performance improvement over the baseline methods.
In particular, by incorporating and learning the class compatibility matrix $\mH$ in an end-to-end fashion, we find that \method{}-Cheby-1 achieves a gain of up to 7\% compared to GCN-Cheby in heterophily settings, while \method{}-MLP-1 performs up to 30\% better in heterophily and 50\% better in homophily compared to the graph-agnostic MLP model.

\vspace{0.1cm}
\noindent \textbf{Real Graphs with Heterophily.} 
Results for graphs with heterophily are presented in Table~\ref{table:real-results-10-features} (top, left).
Overall, we observe that, with respect to mean accuracy, the top-3 methods 
across all the graphs are based on \method{}, 
which demonstrates the importance of incorporating and learning the compatibility matrix $\mH$ into GNNs.
Notably, CPGNN-Cheby-1 
performs the best overall and
significantly outperforms the other baseline methods, achieving 
improvements between 1.68\% and 10.64\% in mean accuracy compared to GNN baselines.
These results demonstrate the effectiveness of \method{} in heterophily settings on real-world benchmarks. 
We note that our empirical analysis also confirms the small time complexity overhead of \method{}: on the Squirrel dataset, the runtimes of \method{}-MLP-1 and \method{}-Cheby-1 are 39s and 697s, respectively, while the prior belief estimators, MLP and GCN-Cheby, run in 29s and 592s in our implementation.

\vspace{0.1cm}
\noindent \textbf{Real-World Graphs with Homophily.}
For the real graphs with homophily, we report the results per method in Table~\ref{table:real-results-10-features} (top, right).
Recall that our framework generalizes GNN for both homophily and heterophily.
We find in Table~\ref{table:real-results-10-features}, the methods from the proposed framework perform better or comparable to the baselines, including those which have an implicit assumption of strong homophily.
Therefore, our methods are more universal while able to maintain the same level of performance as those that are optimized under a strict homophily assumption.
As an aside, we observe that \method{}-Cheby-1 is the best performing method on Pubmed. 

\vspace{0.1cm}
\noindent\textbf{Summary.} 
For the common settings of semi-supervised node classification with contextual features available, the above results show that \method{} variants have the best performance in heterophily settings while maintaining comparable performance in the homophily settings. 
Considering both the heterophily and homophily settings, \method{}-Cheby-1 is the best method overall, which ranked first in the heterophily settings and second in homophily settings.

\subsection{Node Classification \emph{without} Features} \label{sec:exp-classification-without-contextual-features}
Most previous work on semi-supervised node classification has focused only on graphs that have contextual  features on the nodes.
However, %
most graphs do not have such node-level features~\cite{nr}, which greatly limits the utility of the methods proposed in prior work that assume availability of such features. %
Thus, we conduct extensive experiments on semi-supervised node classification without contextual features using the same real graphs as before.

\begin{table*}[t]
\setlength{\tabcolsep}{5pt}
        \centering
        {\small
        \begin{tabular}{l cccc c cccc}
        \toprule
        & \multicolumn{4}{c}{\bf Heterophilous graphs} && \multicolumn{4}{c}{\bf Homophilous graphs} \\ \cmidrule{2-5} \cmidrule{7-10}  
        & \texttt{\bf Texas}  & \texttt{\bf Squirrel} & \texttt{\bf Chameleon} & Mean && \texttt{\bf Citeseer} & \texttt{\bf Pubmed} & \texttt{\bf Cora} & Mean  \\ 
        & {$h$=0.11} & {$h$=0.22} & {$h$=0.23} & Acc && {$h$=0.74} & {$h$=0.8} & {$h$=0.81} & Acc  \\
        \midrule
        
        & \multicolumn{9}{c}{\textbf{With Features}}  \\      
        \cmidrule{2-10}
        
        \textbf{CPGNN-MLP-1} & $63.75{\,\pm\,4.74}$ & $32.70{\,\pm\,1.90}$ & $51.08{\,\pm\,2.29}$ & 49.18 && $71.30{\,\pm\,1.11}$ & $86.40{\,\pm\,0.36}$ & $77.40{\,\pm\,1.10}$ & 78.37\\
        
        \textbf{CPGNN-MLP-2} & \cellcolor{gray!15}$70.42{\,\pm\,2.97}$ & $26.64{\,\pm\,1.23}$ & $55.46{\,\pm\,1.42}$ & 50.84 && $71.48{\,\pm\,1.85}$ & $85.31{\,\pm\,0.70}$ & $81.24{\,\pm\,1.26}$ & 79.34 \\
        
        \textbf{CPGNN-Cheby-1} & $63.13{\,\pm\,5.72}$ & \cellcolor{gray!40}$37.03{\,\pm\,1.23}$ & $53.90{\,\pm\,2.61}$ & \cellcolor{gray!40}51.35 && $72.04{\,\pm\,0.53}$ & \cellcolor{gray!40}$86.68{\,\pm\,0.20}$ & \cellcolor{gray!15}$83.64{\,\pm\,1.31}$ & \cellcolor{gray!15}80.79\\
        
        \textbf{CPGNN-Cheby-2} & $65.97{\,\pm\,8.78}$ & $27.92{\,\pm\,1.53}$ & \cellcolor{gray!40}$56.93{\,\pm\,2.03}$ & 50.27 && \cellcolor{gray!15}$72.06{\,\pm\,0.51}$ & $86.66{\,\pm\,0.24}$ & $81.62{\,\pm\,0.97}$ & 80.11 \\
        
        \cmidrule{1-5} \cmidrule{7-10}
        
        \textbf{\methodbeyond} & \cellcolor{gray!40}$71.39{\,\pm\,2.57}$ & $29.50{\,\pm\,0.77}$ & $48.12{\,\pm\,1.96}$ & \cellcolor{gray!15}49.67 && $71.76{\,\pm\,0.64}$ & $85.93{\,\pm\,0.40}$ & $83.43{\,\pm\,0.95}$ & 80.37 \\

        \textbf{GraphSAGE} & $67.36{\,\pm\,3.05}$ & $34.35{\,\pm\,1.09}$ & $45.45{\,\pm\,1.97}$ & 49.05 && $71.74{\,\pm\,0.66}$ & $85.66{\,\pm\,0.53}$ & $81.60{\,\pm\,1.16}$ & 79.67 \\
        
        \textbf{GCN-Cheby} & $58.96{\,\pm\,3.04}$ & $26.52{\,\pm\,0.92}$ & $36.66{\,\pm\,1.84}$ & 40.71 && $72.04{\,\pm\,0.58}$ & \cellcolor{gray!15}$86.43{\,\pm\,0.31}$ & $83.29{\,\pm\,1.20}$ & 80.58 \\
        
        \textbf{MixHop} &  $62.15{\,\pm\,2.48}$ & \cellcolor{gray!15}$36.42{\,\pm\,3.43}$ & $46.84{\,\pm\,3.47}$ & 48.47 && \cellcolor{gray!40}$73.23{\,\pm\,0.60}$ & $85.12{\,\pm\,0.29}$ & \cellcolor{gray!40}$85.34{\,\pm\,1.23}$ & \cellcolor{gray!40}81.23 
        \\
        
        \textbf{GCN} & $55.90{\,\pm\,2.05}$ & $33.31{\,\pm\,0.89}$ & \cellcolor{gray!15}$52.00{\,\pm\,2.30}$ & 47.07 && $72.27{\,\pm\,0.52}$ & $86.42{\,\pm\,0.27}$ & $83.56{\,\pm\,1.21}$ & 80.75 \\
        
        \textbf{GAT} & $55.83{\,\pm\,0.67}$ & $31.20{\,\pm\,2.57}$ & $50.54{\,\pm\,1.97}$ & 45.86 && $72.63{\,\pm\,0.87}$ & $84.48{\,\pm\,0.22}$ & $79.57{\,\pm\,2.12}$ & 78.89 \\ 
        
        \textbf{MLP} & $64.65{\,\pm\,3.06}$ & $25.50{\,\pm\,0.87}$ & $37.36{\,\pm\,2.05}$ & 42.50 && $66.52{\,\pm\,0.99}$ & $84.70{\,\pm\,0.33}$ & $64.81{\,\pm\,1.20}$ & 72.01 \\ 
        
        \midrule
        & \multicolumn{9}{c}{\textbf{Without Features}}    \\    
        \cmidrule{2-10}

        \textbf{CPGNN-MLP-1} & $64.05{\,\pm\,7.65}$ & \cellcolor{gray!40}$55.19{\,\pm\,1.88}$ & $68.38{\,\pm\,3.48}$ & \cellcolor{gray!40}62.54 && $65.70{\,\pm\,2.96}$ & $81.98{\,\pm\,0.36}$ & $81.97{\,\pm\,1.24}$ & 76.55 \\
        
        \textbf{CPGNN-MLP-2} & $65.14{\,\pm\;9.99}$ & $36.37{\,\pm\,2.08}$ & \cellcolor{gray!40}$70.18{\,\pm\,2.64}$ & 57.23 && $67.66{\,\pm\,2.29}$ & $82.33{\,\pm\,0.39}$ & $82.37{\,\pm\,1.70}$ & 77.46 \\
        
        \textbf{CPGNN-Cheby-1} & $63.78{\,\pm\,7.67}$ & $54.76{\,\pm\,2.01}$ & $67.19{\,\pm\,2.18}$ & 61.91 && \cellcolor{gray!15}$67.93{\,\pm\,2.86}$ & \cellcolor{gray!15}$82.44{\,\pm\,0.58}$ & \cellcolor{gray!15}$83.76{\,\pm\,1.81}$ & \cellcolor{gray!15}78.04 \\
        
        \textbf{CPGNN-Cheby-2} & \cellcolor{gray!40}$70.27{\,\pm\,8.26}$ & $26.42{\,\pm\,1.20}$ & $68.25{\,\pm\,1.57}$ & 54.98 && $67.39{\,\pm\,2.69}$ & $82.27{\,\pm\,0.54}$ & $83.02{\,\pm\,1.29}$ & 77.56 \\
        
        \cmidrule{1-5} \cmidrule{7-10}
        
        \textbf{\methodbeyond} & \cellcolor{gray!15}$68.38{\,\pm\,6.98}$ & 
        \cellcolor{gray!15}$50.91{\,\pm\,1.71}$ & \cellcolor{gray!15}$62.41{\,\pm\,2.14}$ & \cellcolor{gray!15}60.57 && $68.37{\,\pm\,2.93}$ & \cellcolor{gray!40}$82.97{\,\pm\,0.37}$ & $83.22{\,\pm\,1.56}$ & 78.19 \\

        \textbf{GraphSAGE} & $67.03{\,\pm\,4.90}$ & $36.90{\,\pm\,2.36}$ & $58.53{\,\pm\,2.20}$ & 54.15 && $66.71{\,\pm\,3.27}$ & $77.86{\,\pm\,3.84}$ & $81.77{\,\pm\,2.00}$ & 75.45 \\
        
        \textbf{GCN-Cheby} & $50.00{\,\pm\,8.08}$ & $12.62{\,\pm\,0.73}$ & $14.93{\,\pm\,1.53}$ & 25.85 && $67.56{\,\pm\,3.24}$ & $79.14{\,\pm\,0.38}$ & $83.66{\,\pm\,1.02}$ & 76.79 \\
        
        \textbf{MixHop} & $57.57{\,\pm\,5.56}$ & $33.54{\,\pm\,2.08}$ & $50.15{\,\pm\,2.78}$ & 47.08 && $68.38{\,\pm\,3.06}$ & $82.72{\,\pm\,0.75}$ & \cellcolor{gray!40}$84.73{\,\pm\,1.80}$ & \cellcolor{gray!40}78.61 \\
        
        \textbf{GCN} & $51.08{\,\pm\,7.48}$ & $43.78{\,\pm\,1.39}$ & $62.04{\,\pm\,2.17}$ & 52.30 && $67.14{\,\pm\,3.15}$ & $82.28{\,\pm\,0.50}$ & $83.34{\,\pm\,1.38}$ & 77.59 \\
        
        \textbf{GAT} & $57.03{\,\pm\,4.31}$ & $42.46{\,\pm\,2.08}$ & $60.31{\,\pm\,2.61}$ & 53.26 && \cellcolor{gray!40}$68.64{\,\pm\,3.27}$ & $81.92{\,\pm\,0.33}$ & $81.79{\,\pm\,2.21}$ & 77.45 \\ 
        
        \textbf{MLP} & $44.86{\,\pm\,9.29}$ & $19.77{\,\pm\,0.80}$ & $20.57{\,\pm\,2.29}$ & 28.40 && $19.78{\,\pm\,1.35}$ & $39.58{\,\pm\,0.69}$ & $21.61{\,\pm\,1.92}$ & 26.99 \\

        \bottomrule
        \end{tabular}
        }
        \caption{
            Accuracy on graphs   %
            \textit{with} (top) and \textit{without} (bottom) features. Left: heterophilous graphs; Right: homophilous graphs.
        }
        \label{table:real-results-10-features}
\end{table*}

\vspace{0.1cm}
\noindent \textbf{Experimental Setup.} 
To investigate the performance of \method{} and baselines when contextual feature vectors are not available for nodes in the graph, we follow the approach as \citet{kipf2017semi} by replacing the node features $\matX$ in each benchmark with an identity matrix $\matI$. 
We use the training, validation and test splits provided by \citet{pei2019geom}. 

\vspace{0.1cm}
\noindent \textbf{Heterophily.}
We report results on graphs with strong heterophily under the featureless setting in Table~\ref{table:real-results-10-features} (bottom, left). 
We observe that the best performing methods for each dataset are 
all \method{} variants. 
From the 
mean accuracy
perspective, all \method{} variants outperform all baselines except \methodbeyond{}, which is also proposed to handle heterophily, in the overall performance; \method-MLP-1 has the best \emph{overall} performance, followed by \method-Cheby-1. 
It is also worth noting that the performance of GCN-Cheby and MLP, upon which our prior belief estimator is based on, are significantly worse than other methods. 
This demonstrates the effectiveness of incorporating the class compatibility matrix $\mH$ in GNN models and learning it in an end-to-end fashion.

\vspace{0.1cm}
\noindent \textbf{Homophily.} 
We report the results in Table \ref{table:real-results-10-features} (bottom, right). 
The featureless setting for graphs with strong homophily is a fundamentally easier task compared to graphs with strong heterophily, especially for methods with implicit homophily assumptions, as they tend to yield highly similar prediction within the proximity of each node.
Despite this, the \method{} variants  perform comparably to the state-of-the-art methods.

\vspace{0.1cm}
\noindent \textbf{Summary.}
Under the featureless setting, the above results show that \method{} variants achieve state-of-the-art performance in the heterophily graphs, and comparable performance to the best baselines in the homophily graphs. 
Considering both heterophily and homophily, \method{}-Cheby-1 is again the best method overall.

\begin{figure*}[t!]
    \centering
    \begin{subfigure}{0.64\textwidth}
        \centering
        \includegraphics[width=\columnwidth, keepaspectratio]{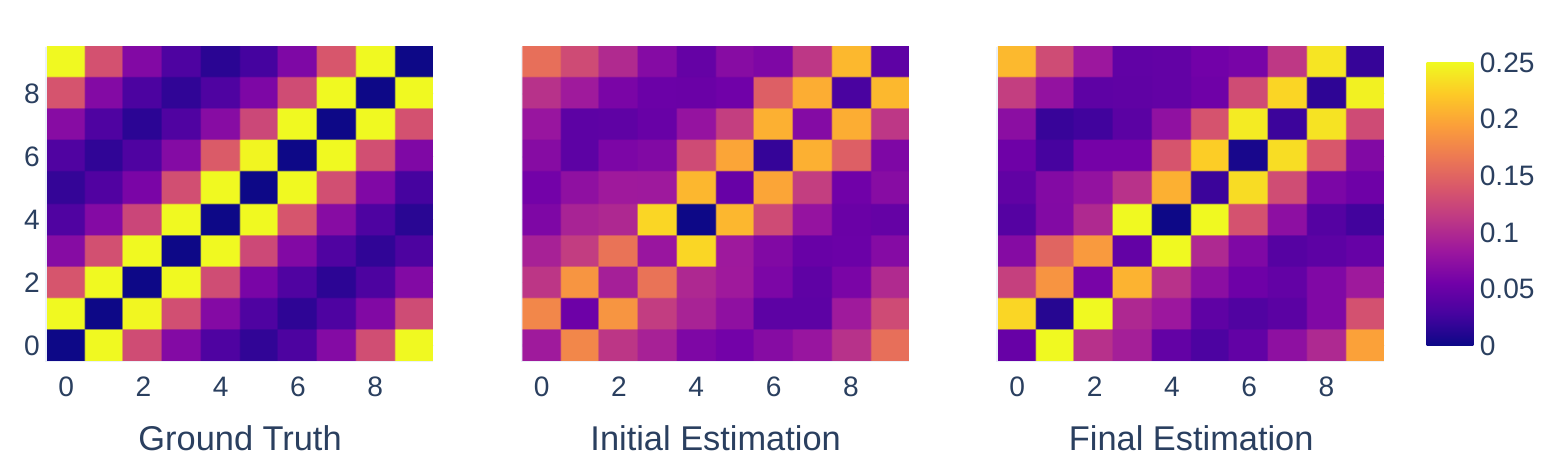}
        \caption{{
            The empirical (ground truth) compatibility matrix, and its initial and final estimations.}}
        \label{fig:heatmap-syn-h}
    \end{subfigure}
    ~
    \begin{subfigure}{0.32\textwidth}
        \centering
        \includegraphics[width=\columnwidth, keepaspectratio, trim={0 0.5cm 1.8cm 2.6cm}, clip]{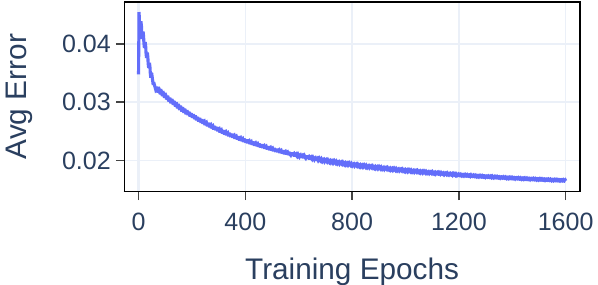}
        \caption{{Error of compatibility matrix estimation throughout the training process.}}
        \label{fig:error-syn-h}
    \end{subfigure}
    \vspace{-1mm}
    \caption{{Visualizations and estimation error of compatibility matrix for an instance of \texttt{syn-products} with $h=0$.}}
    \label{fig:heatmap-error}
    \vspace{-0.4cm}
\end{figure*}

\subsection{Ablation Study}\label{sec:exp-ablation-study}
To evaluate the effectiveness of our model design, we conduct an ablation study by examining variants of \method{}-MLP-1 with one design element removed at a time. Figure~\ref{fig:design-ablations} {in Appendix \S\ref{sec:app-detailed_results}} presents the results for the ablation study, with more detailed results presents in Table \ref{table:exp-ablation-study}. {We also discuss the effectiveness of pretraining in Appendix \S\ref{sec:app-detailed_results}.}

\vspace{0.1cm}
\noindent \textbf{Initialization and Regularization of $\;\mathbf{\bar{H}}$.} Here we study 2 variants of \method{}-MLP-1: (1) No $\mathbf{\bar{H}}$ initialization, when $\mathbf{\bar{H}}$ is initialized using glorot initialization (similar to other GNN formulations) instead of our initialization process described in \S~\ref{sec:training-procedure}. (2) No $\mathbf{\bar{H}}$ regularization, where we remove the regularization term $\Phi(\bar{\mH})$ as defined in Eq. \eqref{eq:H-zero-reg} from the overall loss function (Eq. \eqref{eq:overall-loss}). In Fig. \ref{fig:design-ablation-1}, we see that replacing the initializer can lead to up to 30\% performance drop for the model, while removing the regularization term can cause up to 6\% decrease in performance. These results support our claim that initializing $\mathbf{\bar{H}}$ using pretrained prior beliefs and known labels in the training set and regularizing the $\bar{\mH}$ around 0 lead to better overall performance.

\vspace{0.1cm}
\noindent \textbf{End-to-end Training of $\mathbf{\bar{H}}$.} To demonstrate the performance gain through end-to-end training of \method{} after the initialization of $\mathbf{\bar{H}}$, we compare the final performance of \method{}-MLP-1 with the performance after $\mathbf{\bar{H}}$ is initialized; Fig. \ref{fig:design-ablation-3} shows the results. From the results, we see that the end-to-end training process of \method{} has contributed up to 21\% performance gain. We believe such performance gain is due to a more accurate $\mathbf{\bar{H}}$ 
learned through the training process, as demonstrated in the next subsection. 

\subsection{Compatibility Matrix Estimation} \label{sec:exp-h-estimation}
As described in \S\ref{sec:h-matrix-estimation}, we can obtain an estimation $\hat{\matH}$ of the class compatiblity matrix $\matH \in [0, 1]^{|\setY|\times|\setY|}$ through the learned parameter $\bar{\mathbf{H}}$. To measure the accuracy of the estimation $\hat{\matH}$, we calculate the average error of each element for the estimated $\hat{\matH}$ as following: 
$\bar{\delta}_{\matH} = \frac{\abs{\hat{\matH} - \matH}}{|\setY|^2}$.

Figure \ref{fig:heatmap-error} shows an example of the obtained estimation $\hat{\matH}$ on the synthetic benchmark \texttt{syn-products} with homophily ratio $h=0$ using heatmaps, along with the initial estimation derived following \S\ref{sec:training-procedure} which \method{} optimizes upon, and the ground truth empirical compatibility matrix as defined in Def. \ref{dfn:compatibility-matrix}. From the heatmap, we can visually observe the improvement of the final estimation upon the initial estimation. The curve of the estimation error with respect to the number of training epochs also shows that the estimation error decreases throughout the training process, supporting the observations through the heatmaps. These results illustrate the interpretability of parameters $\bar{\mathbf{H}}$, and effectiveness of our modeling of compatibility matrix.

\section{Related Work} \label{sec:related-work}

\noindent \textbf{SSL before GNNs.} The problem of semi-supervised learning (SSL) %
or \textit{collective classification}~\cite{sen2008collective,mcdowell2007cautious,rossi2012transforming} %
can be solved with 
iterative methods~\cite{NevilleJ00-ICA, QingGetoor03-ICA}, %
graph-based regularization and probabilistic graphical models~\cite{london2014collective}. Among these methods, our approach is related to belief propagation (BP)~\cite{yedidia2003understanding,rossi18-bigdata}, a message-passing approach where each node iteratively sends its neighboring nodes estimations of their beliefs based on its current belief, and updates its own belief based on the estimations received from its neighborhood. 
\citet{koutra2011unifying} and  \citet{gatterbauer2015linearized} have proposed \textit{linearized} versions which are faster to compute. However, these approaches require the class-compatibility matrix to be determined before the inference stage, and cannot support end-to-end training.

\noindent \textbf{GNNs.} In recent years, graph neural networks (GNNs) have become increasingly popular for graph-based semi-supervised node classification problems thanks to their ability to learn through end-to-end training. %
\citet{defferrard2016convolutional} proposed an early version of GNN by generalizing convolutional neural networks (CNNs) from regular grids (e.g., images) to irregular grids (e.g., graphs). %
\citet{kipf2017semi} introduced GCN, a popular GNN model which simplifies the previous work. Other GNN models that have gained wide attention include Planetoid~\cite{yang2016revisiting} and GraphSAGE~\cite{hamilton2017inductive}. 
More recent works have looked into designs which strengthen the effectiveness of GNN to capture graph information: GAT~\cite{velivckovic2017graph} and AGNN~\cite{thekumparampil2018attentionbased} introduced an edge-level attention mechanism; MixHop~\cite{MixHop} and Geom-GCN~\cite{pei2019geom} designed aggregation schemes which go beyond the immediate neighborhood of each node; the jumping knowledge network~\cite{XuLTSKJ18-jkn} leverages representations from intermediate layers; GAM~\cite{gam} and GMNN~\cite{qu2019gmnn} use a separate model to capture the agreement or joint distribution of labels in the graph. 
To capture more graph information, recent works trained very deep networks with 100+ layers~\cite{li2019deepgcns,li2020deepergcn,rong2020dropedge}.

Although many of these GNN methods work well when the data exhibits strong homophily, %
most of them perform poorly in the challenging (and largely overlooked until recently) setting of heterophily. %
Besides Geom-GCN and MixHop that were applied to heterophilous graphs, recent studies 
showed empirically that solutions to oversmoothing can tackle the heterophily setting~\cite{Chen20-heterophily-oversmoothing} and explored the theoretical connections 
of these two problems~\cite{yan2021heterophily-oversmoothing}. 
Also, \citet{zhu2020beyond} discussed designs that improve the representation power of GNNs under heterophily via theoretical and empirical analysis. 
Going beyond these designs that prior GNNs have leveraged, we propose a new GNN model 
 that elegantly combines the powerful notion of compatibility matrix $\mH$ from belief propagation with end-to-end training.

\section{Conclusion} \label{sec:conc}

We propose \method{}, an approach that models an interpretable class compatibility matrix into the GNN framework, and conduct extensive empirical analysis under more realistic settings with fewer training samples and a featureless setup. Through theoretical and empirical analysis, we have shown that the proposed model overcomes the limitations of existing GNN models, especially in the complex settings of heterophily graphs without contextual features. 
\section*{Acknowledgments}
We would like to thank Mark Heimann, Yujun Yan and Leman Akoglu for engaging discussions during the early stage of this work, and the reviewers for their constructive feedback.
This material is based upon work supported by the National Science Foundation under CAREER Grant No.~IIS 1845491, Army Young Investigator Award No.~W911NF1810397, an Adobe Digital Experience research faculty award, an Amazon faculty award, a Google faculty award, and AWS Cloud Credits for Research. We gratefully acknowledge the support of NVIDIA Corporation with the donation of the Quadro P6000 GPU used for this research. 
Any opinions, findings, and conclusions or recommendations expressed in this material are those of the author(s) and do not necessarily reflect the views of the National Science Foundation or other funding parties.

{%
\bibliography{paper} 
}%

\appendix

\clearpage

\section*{Appendix}
\renewcommand{\thesubsection}{\Alph{subsection}}
\numberwithin{table}{subsection}

\subsection{Synthetic Graph Generation}
\label{sec:app-synthetic-graph}

Building upon \citet{MixHop}, we generate synthetic graphs by following a modified preferential attachment process \cite{Barabasi:1999}, which allows us to control the compatibility matrix $\matH$ in the generated graph while keeping a power law degree distribution. We detail the algorithm of synthetic graph generation in Algorithm \ref{algo:syn-graph-generation}. 

\begin{algorithm}[h!]
\caption{Synthetic Graph Generation}
\label{algo:syn-graph-generation}

{\footnotesize 
\SetKwInput{Input}{Input}\SetKwInput{Output}{Output}
\SetKwInput{HyperParams}{Hyper-parameters}\SetKwInput{TrainParams}{Network Parameters}
\Input{%
$C \in \mathbb{N}$: Number of classes in generated graph; \newline
$\mathbf{N} \in \mathbb{N}^{C}$: target size of each class; \newline
$n_0 \in \mathbb{N}$: Number of nodes for the initial bootstrapping graph, which should be much smaller than the total number of nodes; \newline
$m \in \mathbb{N}$: Number of edges added with each new node; \newline
$\matH \in [0,1]^{C\times C}$: Target compatibility matrix for generated graph; \newline
$\graph{G}_r=(\vertexSet_r, \edgeSet_r)$: Reference graph with node set $\vertexSet_r$ and edge set $\edgeSet_r$; \newline
$\vecy_r$: mapping from each node $v \in \vertexSet_r$ to its class label $\vecy_r[v]$ in the reference graph $\graph{G}_r$; \newline
$\V{X}_r$: mapping from each node $v \in \vertexSet_r$ to its node feature vector $\V{X}_r[v]$ in the reference graph $\graph{G}_r$;
}
\Output{%
Generated synthetic graph $\graph{G}=(\vertexSet, \edgeSet)$, with $\vecy: \vertexSet \rightarrow \setY$ as mapping from each node $v \in \vertexSet$ to its class label $\vecy[v]$, and $\V{X}: \vertexSet \rightarrow \mathbb{R}^F$ as mapping from $v \in \vertexSet$ to its node feature vector $\V{X}[v]$.
}
\Begin{
Initialize class label set $\setY \leftarrow \{0, \dots, C-1\}$, node set $\vertexSet \leftarrow \phi$, edge set $\edgeSet \leftarrow \phi$\; 
Calculate the target number of nodes $n$ in the generated graph by summing up all elements in $\mathbf{N}$\;
Generate node label vector $\vecy$, such that class label $y \in \setY$ appears exactly $\mathbf{N}[y]$ times in $\vecy$, and shuffle $\vecy$ randomly after generation\;

\For{$v \in \{0, 1, \dots, n_0-1\}$}{
    Add new node $v$ with class label $\vecy[v]$ into the set of nodes $\vertexSet$\;
    If $v \neq 0$, add new edge $(v-1, v)$ into the set of edges $\edgeSet$\;
    }

\For{$v \in \{n_0, n_0+1, \dots, n-1\}$}{
    Initialize weight vector $\V{w} \leftarrow \V{0}$ and set $\mathcal{T} \leftarrow \phi$\;
    \For{$u \in \vertexSet$}{
        $\V{w}[u] \leftarrow \matH[\vecy[v], \vecy[u]] \times \V{d}[u]$, where $\V{d}[u]$ is the current degree of node $u$\;
        }
    Normalize vector $\V{w}$ such that $\norm{\V{W}}{1} = 1$\;
    Randomly sample $m$ nodes \emph{without replacement} from $\vertexSet$ with probabilities weighted by $\V{w}$, and add the sampled nodes into set $\mathcal{T}$\;
    Add new node $v$ with class label $\vecy[v]$ into the set of nodes $\vertexSet$\;
    \For{$t \in \mathcal{T}$}{
            Add new edge $(t, v)$ into the set of edges $\edgeSet$\;
        }
    }
    Find a valid injection $\Gamma: \vertexSet \rightarrow \vertexSet_r$ such that $\forall u, v \in \vertexSet, \Gamma(u) = \Gamma(v) \Rightarrow u = v$ and $\vecy[u] = \vecy[v] \Leftrightarrow \vecy_r[\Gamma(u)] = \vecy_r[\Gamma(v)]$\;
    \For{$v \in \vertexSet$}{
        $\V{X}[v] \leftarrow \V{X}_r[\Gamma(v)]$\;
        }
  }
}
\end{algorithm}

For the synthetic graph \texttt{syn-products} used in our experiments, we use \texttt{ogbn-products}~\cite{hu2020ogb} as the reference graph $\graph{G}_r$, with parameters $C=10$, $n_0=70$, $m=6$ and the total number of nodes as 10000; all 10 classes share the same size of 1000 nodes. For the compatibility matrix, we set the diagonal elements of $\matH$ to be the same, which we denote as $h$, and we follow the approach in \citet{MixHop} to set the off-diagonal elements.

\begin{table*}[t]
	\centering
 	\resizebox{\textwidth}{!}{
    \small
	\begin{tabular}{lccccccccccc}
	\toprule
	\multirow{2}{*}{\textbf{Methods}} & \multicolumn{11}{c}{\bf Homophily ratio $h$} \\
	\cmidrule(lr){2-12}
    & \textbf{0.00} & \textbf{0.10} & \textbf{0.20} & \textbf{0.30} & \textbf{0.40} & \textbf{0.50} & \textbf{0.60} & \textbf{0.70} & \textbf{0.80} & \textbf{0.90} & \textbf{1.00} \\ 
    
    \midrule
    
    \textbf{\method-MLP-1} & \cellcolor{blue!40}$76.98{\scriptstyle\pm2.01}$ & \cellcolor{blue!40}$67.88{\scriptstyle\pm0.82}$ & \cellcolor{blue!40}$63.21{\scriptstyle\pm1.57}$ & $65.23{\scriptstyle\pm0.85}$ & $74.86{\scriptstyle\pm0.60}$ & $83.25{\scriptstyle\pm1.30}$ & $89.83{\scriptstyle\pm0.16}$ & $94.67{\scriptstyle\pm0.27}$ & $97.13{\scriptstyle\pm0.40}$ & $98.77{\scriptstyle\pm0.27}$ & $99.43{\scriptstyle\pm0.07}$ \\ 
    
    \textbf{\method-MLP-2} & $71.40{\scriptstyle\pm1.90}$ & $65.28{\scriptstyle\pm0.80}$ & $61.76{\scriptstyle\pm1.18}$ & $66.17{\scriptstyle\pm0.40}$ & $74.67{\scriptstyle\pm0.61}$ & $82.63{\scriptstyle\pm0.56}$ & $88.68{\scriptstyle\pm0.21}$ & $93.33{\scriptstyle\pm0.63}$ & $96.81{\scriptstyle\pm0.14}$ & $98.94{\scriptstyle\pm0.18}$ & $99.90{\scriptstyle\pm0.12}$ \\ 

    \textbf{\method-Cheby-1} & $76.37{\scriptstyle\pm0.33}$ & $66.38{\scriptstyle\pm0.39}$ & $63.01{\scriptstyle\pm0.68}$ & \cellcolor{blue!40}$67.39{\scriptstyle\pm0.94}$ & \cellcolor{blue!40}$77.57{\scriptstyle\pm0.39}$ & \cellcolor{blue!40}$86.86{\scriptstyle\pm1.40}$ & \cellcolor{blue!40}$94.44{\scriptstyle\pm0.24}$ & \cellcolor{blue!40}$98.16{\scriptstyle\pm0.16}$ & \cellcolor{blue!40}$99.60{\scriptstyle\pm0.06}$ & \cellcolor{blue!40}$99.89{\scriptstyle\pm0.13}$ & \cellcolor{blue!40}$100.00{\scriptstyle\pm0.00}$ \\ 
    
    \textbf{\method-Cheby-2} & $72.80{\scriptstyle\pm1.72}$ & $65.26{\scriptstyle\pm0.87}$ & $62.05{\scriptstyle\pm1.17}$ & $64.40{\scriptstyle\pm0.86}$ & $73.12{\scriptstyle\pm0.62}$ & $80.90{\scriptstyle\pm1.09}$ & $87.43{\scriptstyle\pm0.22}$ & $91.85{\scriptstyle\pm0.62}$ & $96.66{\scriptstyle\pm0.47}$ & $99.67{\scriptstyle\pm0.09}$ & \cellcolor{blue!40}$100.00{\scriptstyle\pm0.00}$ \\ 
    
    \midrule
    \textbf{GraphSAGE} & $59.15{\scriptstyle\pm0.73}$ & $53.53{\scriptstyle\pm0.77}$ & $54.54{\scriptstyle\pm0.66}$ & $56.08{\scriptstyle\pm0.45}$ & $61.17{\scriptstyle\pm1.19}$ & $68.98{\scriptstyle\pm1.44}$ & $78.14{\scriptstyle\pm1.10}$ & $86.55{\scriptstyle\pm0.30}$ & $92.71{\scriptstyle\pm1.35}$ & $96.69{\scriptstyle\pm0.38}$ & $99.12{\scriptstyle\pm0.11}$ \\ 
    
    \textbf{GCN-Cheby} & $68.65{\scriptstyle\pm1.30}$ & $60.51{\scriptstyle\pm1.64}$ & $61.98{\scriptstyle\pm0.68}$ & $66.20{\scriptstyle\pm1.24}$ & $74.43{\scriptstyle\pm1.40}$ & $83.60{\scriptstyle\pm0.77}$ & $92.28{\scriptstyle\pm0.47}$ & $97.11{\scriptstyle\pm0.18}$ & $99.25{\scriptstyle\pm0.09}$ & $99.81{\scriptstyle\pm0.06}$ & $99.80{\scriptstyle\pm0.18}$ \\ 
    
    \textbf{MixHop} & $10.66{\scriptstyle\pm1.73}$ & $11.29{\scriptstyle\pm0.61}$ & $12.40{\scriptstyle\pm1.86}$ & $11.87{\scriptstyle\pm2.38}$ & $13.91{\scriptstyle\pm3.21}$ & $19.72{\scriptstyle\pm1.06}$ & $20.33{\scriptstyle\pm1.11}$ & $21.72{\scriptstyle\pm2.07}$ & $21.88{\scriptstyle\pm3.04}$ & $21.32{\scriptstyle\pm1.91}$ & $22.60{\scriptstyle\pm2.14}$ \\ 
    
    \textbf{GCN} & $44.72{\scriptstyle\pm0.51}$ & $41.87{\scriptstyle\pm1.37}$ & $46.49{\scriptstyle\pm0.50}$ & $55.63{\scriptstyle\pm0.88}$ & $69.33{\scriptstyle\pm0.80}$ & $81.21{\scriptstyle\pm0.97}$ & $90.65{\scriptstyle\pm0.35}$ & $96.01{\scriptstyle\pm0.10}$ & $98.80{\scriptstyle\pm0.14}$ & $99.64{\scriptstyle\pm0.01}$ & $99.99{\scriptstyle\pm0.01}$ \\
    
    \textbf{GAT} & $19.59{\scriptstyle\pm5.96}$ & $21.74{\scriptstyle\pm2.06}$ & $25.67{\scriptstyle\pm1.77}$ & $30.34{\scriptstyle\pm2.90}$ & $39.42{\scriptstyle\pm7.60}$ & $50.62{\scriptstyle\pm5.45}$ & $64.68{\scriptstyle\pm5.01}$ & $88.01{\scriptstyle\pm3.71}$ & $98.01{\scriptstyle\pm0.65}$ & $99.06{\scriptstyle\pm0.80}$ & $99.94{\scriptstyle\pm0.02}$ \\
    
    \textbf{MLP} & $47.46{\scriptstyle\pm2.66}$ & $47.15{\scriptstyle\pm1.47}$ & $47.55{\scriptstyle\pm0.90}$ & $47.35{\scriptstyle\pm2.02}$ & $47.07{\scriptstyle\pm0.94}$ & $48.25{\scriptstyle\pm0.76}$ & $47.37{\scriptstyle\pm1.41}$ & $47.38{\scriptstyle\pm1.64}$ & $46.87{\scriptstyle\pm0.65}$ & $46.94{\scriptstyle\pm0.86}$ & $48.12{\scriptstyle\pm1.63}$ \\ 
    
    \bottomrule
	\end{tabular}
 	}
     \caption{Node classification \emph{with} features on synthetic graph \texttt{syn-products}  (\S\ref{sec:exp-classification-with-contextual-features}, Fig.~\ref{fig:syn-benchmark-results}): mean classification accuracy and standard deviation per method for different values of homophily ratio $h$.} %
     \label{table:syn-products-table-10-est-h}
\end{table*}

\begin{table*}[t]
	\centering
 	\resizebox{\textwidth}{!}{
    \small
	\begin{tabular}{lccccccccccc}
	\toprule
	\multirow{2}{*}{\textbf{Variants}} & \multicolumn{11}{c}{\bf Homophily ratio $h$} \\
	\cmidrule(lr){2-12}
	& \textbf{0.00} & \textbf{0.10} & \textbf{0.20} & \textbf{0.30} & \textbf{0.40} & \textbf{0.50} & \textbf{0.60} & \textbf{0.70} & \textbf{0.80} & \textbf{0.90} & \textbf{1.00} \\ 
    \midrule
    \textbf{\method-MLP-1} & $77.52{\scriptstyle\pm1.82}$ & $67.95{\scriptstyle\pm0.68}$ & $62.55{\scriptstyle\pm1.73}$ & $64.85{\scriptstyle\pm0.81}$ & $74.67{\scriptstyle\pm0.82}$ & $82.95{\scriptstyle\pm1.07}$ & $89.75{\scriptstyle\pm0.24}$ & $94.64{\scriptstyle\pm0.38}$ & $97.17{\scriptstyle\pm0.34}$ & $98.73{\scriptstyle\pm0.30}$ & $99.35{\scriptstyle\pm0.10}$ \\ 
    \midrule
    \textbf{No $\bar{\mathbf{H}}$ Init. (Fig. \ref{fig:design-ablation-1})} & $46.74{\scriptstyle\pm2.86}$ & $46.03{\scriptstyle\pm1.56}$ & $46.20{\scriptstyle\pm0.61}$ & $46.53{\scriptstyle\pm1.55}$ & $47.20{\scriptstyle\pm0.94}$ & $54.54{\scriptstyle\pm0.50}$ & $67.47{\scriptstyle\pm2.06}$ & $80.18{\scriptstyle\pm0.53}$ & $87.44{\scriptstyle\pm1.93}$ & $91.89{\scriptstyle\pm1.59}$ & $95.47{\scriptstyle\pm0.68}$ \\ 
    
    \textbf{No $\bar{\mathbf{H}}$ Reg. (Fig. \ref{fig:design-ablation-1})} & $71.49{\scriptstyle\pm2.01}$ & $61.88{\scriptstyle\pm1.37}$ & $58.98{\scriptstyle\pm0.64}$ & $60.43{\scriptstyle\pm1.82}$ & $68.91{\scriptstyle\pm1.35}$ & $78.55{\scriptstyle\pm1.14}$ & $85.47{\scriptstyle\pm1.99}$ & $91.76{\scriptstyle\pm0.49}$ & $95.48{\scriptstyle\pm0.59}$ & $97.31{\scriptstyle\pm0.03}$ & $98.82{\scriptstyle\pm0.19}$ \\ 
    
    \textbf{After $\bar{\mathbf{H}}$ Init. (Fig. \ref{fig:design-ablation-3})} & $56.49{\scriptstyle\pm4.48}$ & $52.22{\scriptstyle\pm2.70}$ & $51.02{\scriptstyle\pm1.08}$ & $53.05{\scriptstyle\pm2.25}$ & $57.58{\scriptstyle\pm1.99}$ & $62.95{\scriptstyle\pm1.51}$ & $68.31{\scriptstyle\pm1.88}$ & $72.18{\scriptstyle\pm1.90}$ & $77.53{\scriptstyle\pm1.56}$ & $82.13{\scriptstyle\pm1.11}$ & $88.38{\scriptstyle\pm1.77}$ \\

    \textbf{No Pretrain (Fig. \ref{fig:design-ablations-extra})} & $75.67{\scriptstyle\pm2.65}$ & $65.45{\scriptstyle\pm0.47}$ & $60.23{\scriptstyle\pm0.68}$ & $64.15{\scriptstyle\pm0.97}$ & $73.59{\scriptstyle\pm0.79}$ & $82.83{\scriptstyle\pm0.52}$ & $88.92{\scriptstyle\pm0.60}$ & $94.81{\scriptstyle\pm0.79}$ & $97.36{\scriptstyle\pm0.09}$ & $98.97{\scriptstyle\pm0.18}$ & $99.52{\scriptstyle\pm0.08}$ \\

    \bottomrule
	\end{tabular}
 	}
     \caption{Ablation study for \method-MLP-1 on \texttt{syn-products} (\S\ref{sec:exp-ablation-study} and App. \S\ref{sec:app-detailed_results}): mean classification accuracy and standard deviation per method for different values of homophily ratio $h$.}
     \label{table:exp-ablation-study}
\end{table*}

\subsection{More Experimental Setups}
\label{sec:app-experimental-specs}

\paragraph{Baseline Implementations.} For all GNN baselines, we use the official GitHub implementations released by the authors:
\begin{itemize}
    \item \textbf{GCN \& GCN-Cheby}~\cite{kipf2017semi}: \url{https://github.com/tkipf/gcn}
    \item \textbf{GraphSAGE}~\cite{hamilton2017inductive}: \url{https://github.com/williamleif/graphsage-simple} (PyTorch implementation)
    \item \textbf{MixHop}~\cite{MixHop}: \url{https://github.com/samihaija/mixhop}
    \item \textbf{GAT}~\cite{velivckovic2017graph}: \url{https://github.com/PetarV-/GAT}
    \item \textbf{\methodbeyond}~\cite{zhu2020beyond}: \url{https://github.com/GemsLab/H2GCN}
    \item \textbf{MLP}: We utilize the \methodbeyond{} implementation, which runs MLP when the \texttt{network\_setup} argument is set to \texttt{M64-R-MO}.
\end{itemize}

\paragraph{\method Implementation.}
The code for our proposed framework, \method, can be found at {\url{https://github.com/GemsLab/CPGNN}}. 

\paragraph{Hardware and Software Specifications.} We ran all the experiments on a workstation which features an AMD Ryzen 9 3900X CPU with 12 cores, 64GB RAM, a Nvidia Quadro P6000 GPU with 24GB GPU Memory and a Ubuntu 20.04.1 LTS operating system. We implemented \method{} using TensorFlow 2.2 with GPU support.

\subsection{Hyperparameter Tuning}
\label{sec:app-hyperparameter}

Below we list the hyperparameters tested per model on each real-world benchmark. %
As the hyperparameters defined by each baseline model differ significantly, we list the combinations of non-default command line arguments that we tested, without explaining them in detail. For further details on the arguments and their definitions, we refer the interested reader to the corresponding original implementations. When multiple hyperparameters are listed, the results reported for each benchmark are based on the hyperparameters which yield the best validation accuracy on average. 

To ensure a fair evaluation of the performance improvement brought by \method{}, the MLP and GCN-Cheby prior belief estimator in \method-MLP and \method-Cheby share the same network architecture (including numbers and sizes of hidden layers) as our MLP and GCN-Cheby baselines. 

\begin{itemize}
    \item \textbf{GraphSAGE}~\cite{hamilton2017inductive}: 
    \begin{itemize}
        \item \texttt{\lstinline{hid_units}}: 64
        \item \texttt{\lstinline{lr}}: $a \in \{0.1, 0.7\}$
        \item \texttt{\lstinline{epochs}}: 500
    \end{itemize}
    
    \item \textbf{GCN-Cheby}~\cite{kipf2017semi}: 
    \begin{itemize}
        \item \texttt{\lstinline{hidden1}}: 64
        \item \texttt{\lstinline{weight_decay}}: $a \in \{\texttt{1e-5, 5e-4}\}$
        \item \texttt{\lstinline{max_degree}}: 2
        \item \texttt{\lstinline{early_stopping}}: 40
    \end{itemize}
    
    \item \textbf{Mixhop}~\cite{MixHop}:
    \begin{itemize}
        \item \texttt{\lstinline{adj_pows}}: 0, 1, 2
        \item \texttt{\lstinline{hidden_dims_csv}}: 64
    \end{itemize}
    \item \textbf{GCN}~\cite{kipf2017semi}: 
    \begin{itemize}
        \item \texttt{\lstinline{hidden1}}: 64
        \item \texttt{\lstinline{early_stopping}}: $a \in \{40, 100, 200\}$
        \item \texttt{\lstinline{epochs}}: 2000
    \end{itemize}
    
    \item \textbf{GAT}~\cite{velivckovic2017graph}: 
    \begin{itemize}
        \item \texttt{\lstinline{hid_units}}: $8$
        \item \texttt{\lstinline{n_heads}}: $8$ 
    \end{itemize}
    
    \item \textbf{\methodbeyond}~\cite{zhu2020beyond}:
    \begin{itemize}
        \item \texttt{\lstinline{network_setup}}: \\
        {\ttfamily M64-T1-G-V-T2-G-V-C1-C2-D-MO} \\
        or {\ttfamily M64-R-T1-G-V-T2-G-V-C1-C2-D-MO} \\
        (\methodbeyond-2 with or without \texttt{ReLU})
        \item \texttt{\lstinline{dropout}}: $0$ or $0.5$
        \item \texttt{\lstinline{l2_regularize_weight}}: \texttt{1e-5}
    \end{itemize}

    \item \textbf{MLP}: 
    \begin{itemize}
        \item Dimension of Feature Embedding: 64
        \item Number of hidden layer: 1
        \item Non-linearity Function: \texttt{ReLU}
        \item Dropout Rate: 0
    \end{itemize}
\end{itemize}

\begin{figure*}[t!]
    \centering
    \begin{subfigure}{0.3\linewidth}
        \centering
        \includegraphics[keepaspectratio, width=.97\textwidth,trim={0 0 2cm 2.6cm},clip]{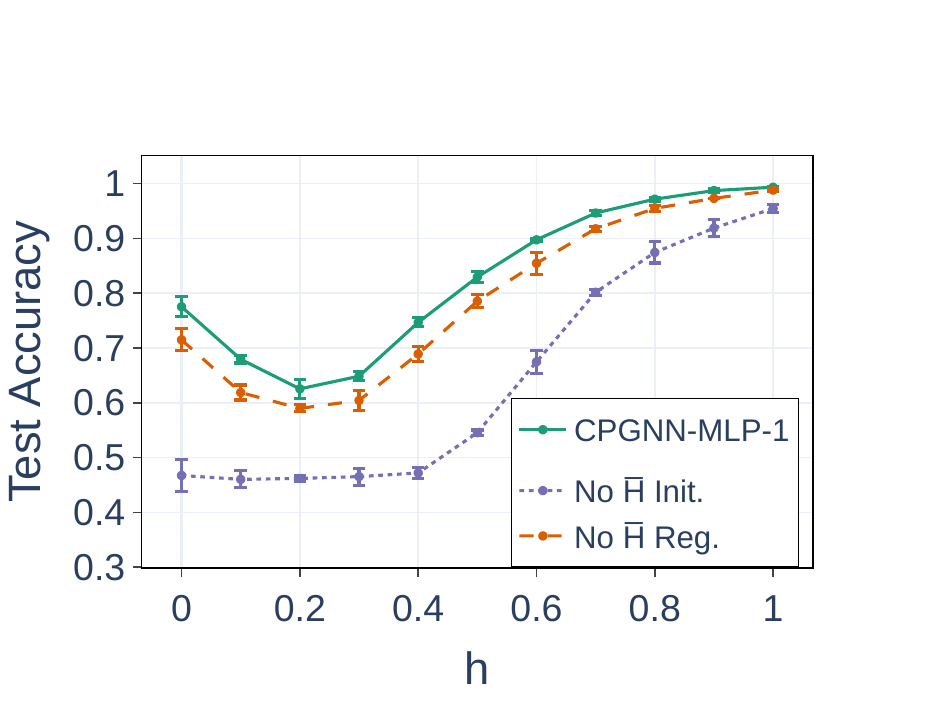}
        \vspace{-2mm}
        \caption{Accuracy without initialization or regularization of $\bar{\mathbf{H}}$.}%
        \label{fig:design-ablation-1}
    \end{subfigure}
    ~
    \begin{subfigure}{0.3\linewidth}
        \centering
        \includegraphics[keepaspectratio, width=.97\textwidth,trim={0 0 2cm 2.6cm},clip]{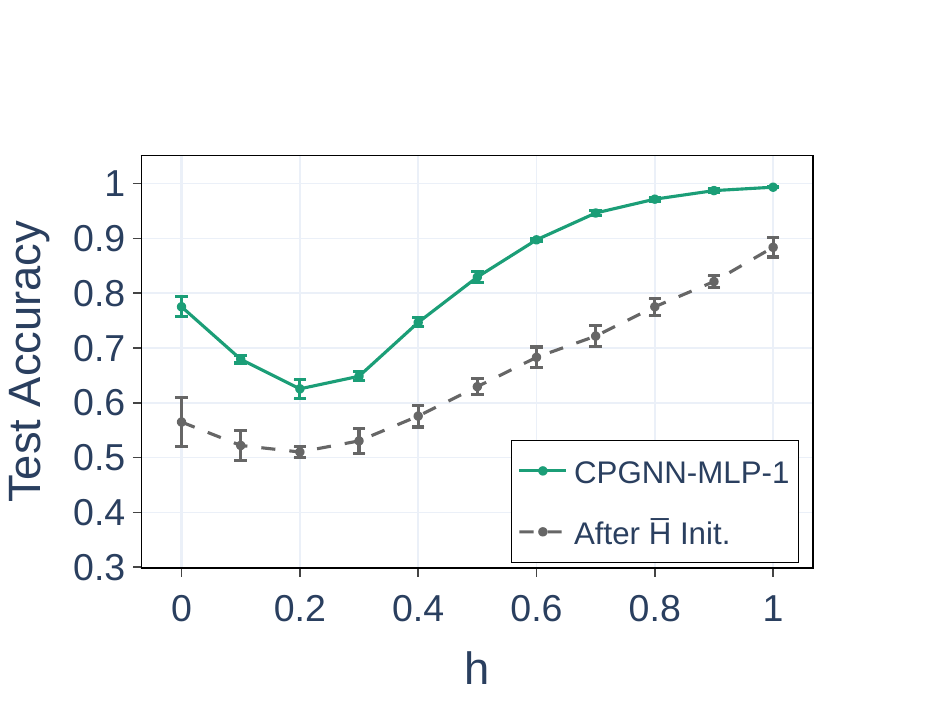}
        \vspace{-2mm}
        \caption{Accuracy after end-to-end training vs. after initializing $\bar{\mathbf{H}}$. } %
        \label{fig:design-ablation-3}
    \end{subfigure}
    \begin{subfigure}{0.32\linewidth}
        \centering
        \includegraphics[keepaspectratio, width=0.9\columnwidth,trim={0 0 2cm 2.6cm},clip]{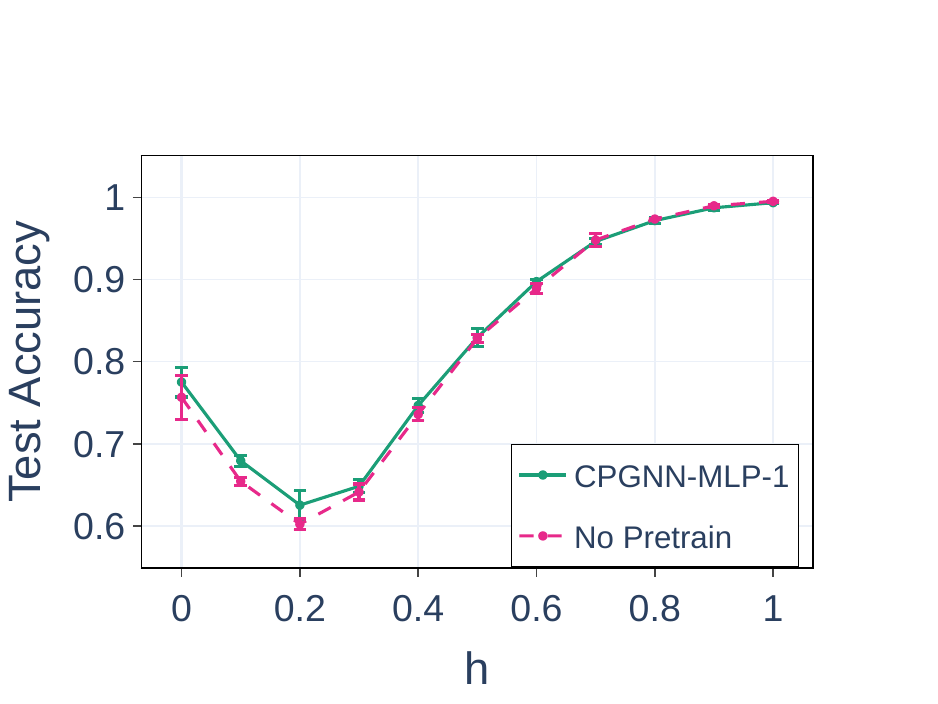}
        \vspace{-2mm}
        \caption{Accuracy with and without pretraining of the prior belief estimator.}
        \label{fig:design-ablations-extra}
    \end{subfigure}
    \caption{Ablation Study: Mean accuracy as a function of $h$. (\subref{fig:design-ablation-1}): When replacing $\bar{\mathbf{H}}$ initialization with glorot or removing $\bar{\mathbf{H}}$ regularization, the performance of \method{} drops significantly;
    (\subref{fig:design-ablation-3}): The significant increase in performance shows the effectiveness of the end-to-end training in our framework;
    (\subref{fig:design-ablations-extra}): {Pretraining contributes up to 2\% performance gain (cf. App. \S\ref{sec:app-detailed_results}).}
    }
    \label{fig:design-ablations}
    \vspace{-0.3cm}
\end{figure*}

\subsection{Detailed Results}\label{sec:app-detailed_results}

\paragraph{Node Classification with Contextual Features.} Table \ref{table:syn-products-table-10-est-h} provides the detailed results on \texttt{syn-products}, which are illustrated in Fig.~\ref{fig:syn-benchmark-results} in \S\ref{sec:exp-classification-with-contextual-features}.

\subsubsection{Ablation Study.} Table \ref{table:exp-ablation-study} presents more detailed results for the ablation study (cf. \S\ref{sec:exp-ablation-study}), complementing Fig. \ref{fig:design-ablations}.
In addition, we conduct an ablation study to examine the effectiveness of {pretraining by 
testing a variant where we skip the pretraining for the prior belief estimator. Figure~\ref{fig:design-ablations-extra} and Table \ref{table:exp-ablation-study} reveal that, though the differences in performance are small, the adoption of pretraining results in up to 2\% increase in performance under heterophily. }

\clearpage

\end{document}